\documentclass[conference]{IEEEtran}
\usepackage{times}
\usepackage[numbers]{natbib}
\usepackage{multicol}
\usepackage[bookmarks=true]{hyperref}
\usepackage{graphicx} 
\usepackage{relsize}
\usepackage[normalem]{ulem}

\usepackage{multirow}
\usepackage{multicol}
\usepackage{graphicx}
\usepackage{xspace}
\usepackage{xcolor}
\usepackage{caption}
\usepackage{wrapfig}
\usepackage{bbding}  
\usepackage{pifont}  

\usepackage{booktabs}
\usepackage{float}
\usepackage{amsmath}  
\usepackage{amssymb}  
\usepackage{bm}
\usepackage{upgreek}

\usepackage{dsfont}
\newcommand{\ours}[0]{\textsc{HUSKY}\xspace}

\usepackage{hyperref}
\usepackage[capitalise, nameinlink]{cleveref}

\usepackage{colortbl}
\definecolor{ourcolor}{HTML}{99e0eb}
\definecolor{ourblue}{HTML}{27a2c3}

\definecolor{tablecolor}{HTML}{ccf2f5} 

\definecolor{tablecolor2}{HTML}{ffcdb4}
\definecolor{citecolor}{HTML}{fe7b5b}
\definecolor{grey}{rgb}{0.9, 0.9, 0.9}
\usepackage{amssymb}

\usepackage{listings}
\definecolor{gred}{rgb}{0.859,0.267,0.216}
\definecolor{ggreen}{rgb}{0.059,0.616,0.345}

\definecolor{deepblue}{HTML}{27a2c3}

\definecolor{deepred}{HTML}{fe7b5b}

\usepackage[font=small,labelfont=bf]{caption}

\usepackage[font=footnotesize,labelfont=bf]{caption}

\definecolor{citecolor}{HTML}{faa700} 
\definecolor{lblue}{HTML}{ffb114} 
\definecolor{ogreen}{HTML}{2E7D32}
\definecolor{bred}{HTML}{BF360C}
\definecolor{newbrown}{HTML}{795548}

\definecolor{lightblue}{HTML}{1399b2}

\hypersetup{
    colorlinks=True,
    linkcolor=lightblue,
    filecolor=magenta,      
    urlcolor=lightblue,
    citecolor=lightblue,
}

\usepackage{multicol}
\usepackage{multirow}
\usepackage{colortbl}
\usepackage{booktabs}   
\usepackage{bbding} 
\usepackage{graphicx}
\let\labelindent\relax 
\usepackage{enumitem}

\usepackage{bbding}

\usepackage{pifont}
\usepackage{xfrac}
\usepackage{arydshln}

\usepackage{tabularx}

\begin{document}
\title{\textsc{HUSKY}: Humanoid Skateboarding System via Physics-Aware Whole-Body Control}

\author{\authorblockN{Jinrui Han\textsuperscript{1,2*} \quad Dewei Wang\textsuperscript{1,3*}
\quad Chenyun Zhang\textsuperscript{1}
\quad Xinzhe Liu\textsuperscript{1,4} \quad \\
Ping Luo\textsuperscript{2} \quad Chenjia Bai\textsuperscript{1\dag{}} \quad Xuelong Li\textsuperscript{1\dag{}} 
}
\authorblockA{
\textsuperscript{1}Institute of Artificial Intelligence (TeleAI), China Telecom \quad 
\textsuperscript{2}The University of Hong Kong \quad \\
\textsuperscript{3}University of Science and Technology of China \quad 
\textsuperscript{4}ShanghaiTech University \quad   \\
\textsuperscript{$*$}Equal contribution \quad
\textsuperscript{\dag{}}Corresponding author\\
Webpage: \href{https://husky-humanoid.github.io/}{\texttt{husky-humanoid.github.io}} \quad Code: \href{https://github.com/TeleHuman/humanoid_skateboarding}{\texttt{TeleHuman/humanoid\_skateboarding}}
} 
}

\twocolumn[{%
\renewcommand\twocolumn[1][]{#1}%
\maketitle
\vspace{-0.45cm}
\begin{center}
    \centering
    \captionsetup{type=figure}
     \includegraphics[width=1.0\textwidth]{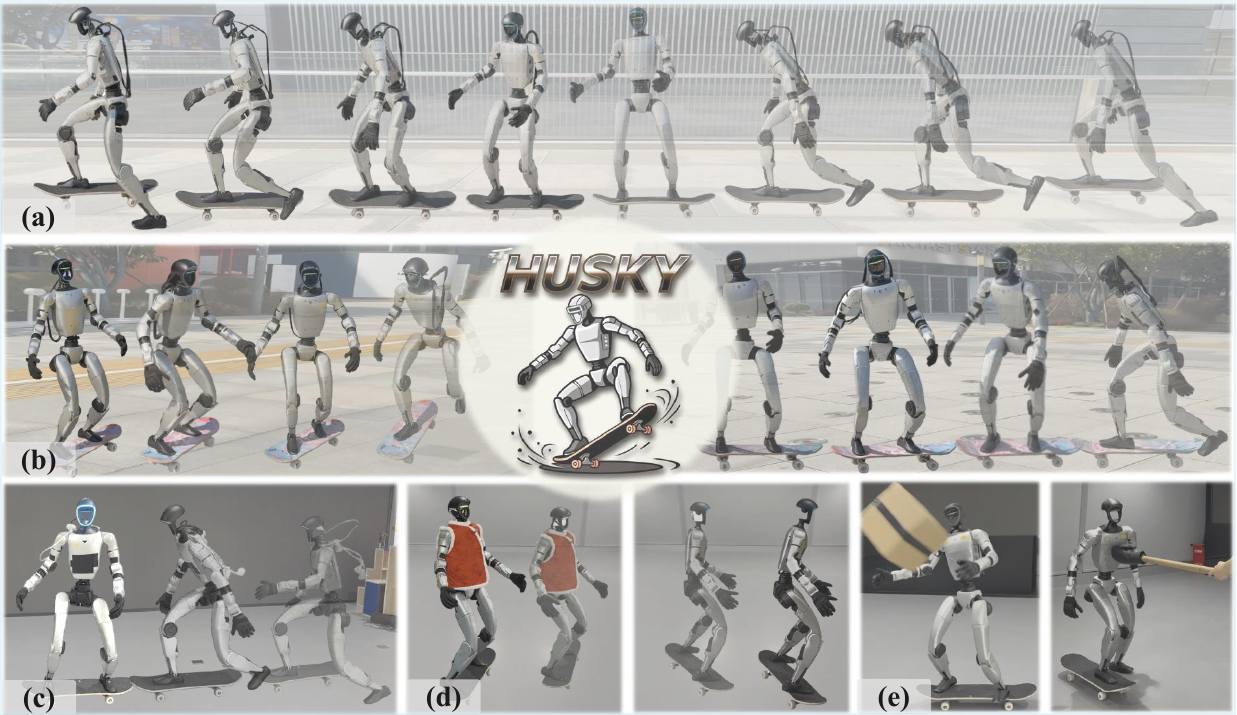 }
     \vspace{-0.17in}
    \caption{\textbf{Overview.} (a) Our proposed framework \ours enables the humanoid robot to perform complete real-world skateboarding, including pushing, steering, and phase transitions. (b) Generalization to diverse outdoor scenarios and skateboards with consistent stability and control. (c) Reliable indoor skateboarding performance. (d) Lean-to-steer behaviors achieved by exploiting robot body tilt. (e) Robustness against external disturbances.} 
    \label{fig:cover}
\end{center}
\vspace{0.04in}
}]

\begin{abstract}
While current humanoid whole-body control frameworks predominantly rely on the static environment assumptions, addressing tasks characterized by high dynamism and complex interactions presents a formidable challenge.
In this paper, we address humanoid skateboarding, a highly challenging task requiring stable dynamic maneuvering on an underactuated wheeled platform. This integrated system is governed by non-holonomic constraints and tightly coupled human-object interactions. Successfully executing this task requires simultaneous mastery of hybrid contact dynamics and robust balance control on a mechanically coupled, dynamically unstable skateboard.
To overcome the aforementioned challenges, we propose \ours{}, a learning-based framework that integrates humanoid-skateboard system modeling and physics-aware whole-body control. 
We first model the coupling relationship between board tilt and truck steering angles, enabling a principled analysis of system dynamics.
Building upon this, \ours{} leverages Adversarial Motion Priors (AMP) to learn human-like pushing motions and employs a physics-guided, heading-oriented strategy for lean-to-steer behaviors. Moreover, a trajectory-guided mechanism ensures smooth and stable transitions between pushing and steering.
Experimental results on the Unitree G1 humanoid platform demonstrate that our framework enables stable and agile maneuvering on skateboards in real-world scenarios. More videos and code are available on our \href{https://husky-humanoid.github.io/}{project page}.

\end{abstract}

\IEEEpeerreviewmaketitle

\section{Introduction}

Recent advancements in robotic hardware and control algorithms have enabled humanoid robots to perform a wide range of whole-body control tasks, including locomotion over complex terrains~\cite{hl:radosavovic2024real, hl:wang2025more, hl:long2025learning}, dancing~\cite{he2025asap, hm:liao2025beyondmimic,xie2026kungfubot,hm:han2025kungfubot2}, manipulation~\cite{homie, shi2026adversarial, song2025hume}, and fall recovery~\cite{hr:huang2025learning, hr:jeong2016efficient}.
However, learning skateboarding skills for humanoid robots remains a highly challenging and largely unexplored problem.
While existing approaches primarily focus on contact-rich whole-body control tasks~\cite{zhang2024wococolearningwholebodyhumanoid, hm:yang2025omniretarget} and humanoid–object interaction scenarios~\cite{physhsi, hdmi}, humanoid skateboarding involves even more intricate contact dynamics and hybrid motion transitions.
Furthermore, the shift from interacting with static objects to an underactuated wheeled platform challenges existing control frameworks, rendering them inadequate for stable humanoid skateboarding system.

Traditional model-based approaches typically leverage simplified dynamic models for trajectory generation and Model Predictive Control (MPC) to synthesize skateboarding maneuvers for legged robots~\cite{takasugi2016real, takasugi2018extended, xu2024optimization}. While these classical methodologies can generate controlled motions, the high computational cost of solving high-dimensional, non-convex optimization problems often precludes the real-time responsiveness required for dynamic skateboarding. Moreover, simplified models are insufficient to capture the complex non-holonomic and underactuated dynamics inherent to skateboarding, resulting in limited robustness to unmodeled physical effects and environmental variations. Although agile skateboarding skills can be achieved in simulation~\cite{mikkola2024online}, they overlook physical hardware constraints and the sim-to-real gap, making direct real-world deployment highly challenging.

Deep Reinforcement Learning (DRL) has recently emerged as a powerful alternative, leveraging large-scale parallel training to synthesize complex behaviors with minimal modeling assumptions and robust performance. 
Recent researches have demonstrated the effectiveness of RL in diverse tasks, from force-controlled interaction~\cite{zhang2025falcon} and end-effector control~\cite{li2025holdbeerlearninggentle} to quadrupedal skateboarding~\cite{liu2025discrete, belov2025quadrupedal} and humanoid roller skating~\cite{gu2026skater}, highlighting RL's capacity to manage complex contact dynamics and stabilize high-speed, underactuated systems.

However, humanoid skateboarding introduces substantially greater challenges due to the emergence of a tightly coupled, highly underactuated system. While the humanoid’s high-dimensional state and action spaces already complicate whole-body control, this difficulty is compounded by the dynamic nature of the skateboard. The board's motion, governed by intricate interactions between the wheels, ground, and the robot’s feet, creates a scenario where the robot must indirectly control its own moving support base.
The inherent instability of this task becomes even more evident when compared to quadrupedal platforms. Unlike quadrupeds~\cite{liu2025discrete}, humanoids possess significantly higher Degrees of Freedom (DoF) and an elevated Center of Mass (CoM), paired with a much smaller and more precarious support polygon. These physical constraints, combined with the need for complex leg reorientation and side-on steering maneuvers, result in highly nonlinear dynamics and severe sim-to-real discrepancies. Consequently, existing learning-based controllers that may succeed on simpler platforms struggle to maintain robust coordination here. Table~\ref{tab:humanoid_vs_quadruped} further highlights these unique structural and task-specific difficulties.

\begin{table}[ht]
\centering
\caption{Comparison of Humanoid and Quadrupedal Skateboarding Tasks. }
\label{tab:humanoid_vs_quadruped}
\begin{tabular}{lcc}
\toprule
\textbf{Term} & \textbf{Humanoid (G1)} & \textbf{Quadruped (Go1)}~\cite{liu2025discrete} \\
\midrule

Degrees of Freedom & 23 & 12 \\
Contact Points & 1--2 & 2--4\\
CoM Height & 0.6--0.8 $m$& 0.2--0.3 $m$ \\
Support Polygon & 0.05--0.10 $m^2$& 0.10--0.15 $m^2$ \\
Side-on Steering & $\checkmark$ & $\times$  \\
Leg Reorientation & $\checkmark$ & $\times$  \\

\bottomrule
\end{tabular}
\end{table}

In this paper, we propose \ours{}, a physics-aware whole-body controller for \uline{\textbf{hu}}manoid \uline{\textbf{sk}}ateboarding s\uline{\textbf{y}}stem.
\ours{} first explicitly analyzes the coupled humanoid–skateboard system and derives an equality constraint between board tilt and truck steering angles, enabling physics-informed policy learning.
The skateboarding task is formulated as a hybrid dynamical system with discrete contact phases, capturing the distinct pushing and steering behaviors inherent to skateboard locomotion. 
Based on this modeling, we employ DRL to train the humanoid to acquire the essential skateboarding skills, including propulsion through pushing and directional control through steering.
To enhance the robustness and naturalness of pushing behaviors, \ours{} leverages Adversarial Motion Priors (AMP)~\cite{peng2021amp} to guide the policy toward human-like propulsion motions.
Steering is achieved through a physics-aware strategy that exploits the intrinsic coupling between humanoid body lean and skateboard truck steering. To facilitate exploration across distinct phases and ensure smooth transitions, \ours{} integrates a trajectory planning mechanism that stabilizes transitions between pushing and steering, resulting in continuous and coordinated skateboarding behaviors.

Taken together, \ours{} enables the humanoid to achieve agile, stable skateboarding maneuvers, as validated in both simulation and real-world experiments.
Our main contributions are summarized as follows:

\begin{itemize}
    \item We model the humanoid skateboarding system by explicitly incorporating the tilt–steering constraint and formulate the skateboarding task as a hybrid dynamical system characterized by distinct contact phases.
    \item We develop a learning-based whole-body control framework that integrates AMP-based pushing, physics-guided steering, and trajectory planning for phase transitions.
    \item We validate \ours{} in both simulation and real-world experiments, demonstrating agile, stable, and human-like humanoid skateboarding maneuvers.
\end{itemize}

\section{Related Work}

\subsection{Learning-based Humanoid Whole-Body Control}
Recent advances in learning-based humanoid whole-body control have enabled robots to tackle complex control tasks, leveraging state-of-the-art algorithms and high-fidelity simulators~\cite{zakka2026mjlablightweightframeworkgpuaccelerated, mittal2025isaac, makoviychuk2021isaac}. Existing research primarily focuses on locomotion~\cite{hl:radosavovic2024real, gu2024humanoid}, motion tracking~\cite{peng2018deepmimic,he2025asap}, and teleoperation~\cite{he2024omniho, ht:ze2025twist,ze2025twist2scalableportableholistic}. These approaches have achieved robust blind locomotion over uneven terrains~\cite{gu2024advancing, xie2025humanoid}, highly dynamic motion tracking~\cite{xie2026kungfubot, hm:han2025kungfubot2}, agile parkour~\cite{zhuang2024humanoid}, and complex manipulation tasks~\cite{fu2024humanplus}. To enhance environmental perception and reactive capabilities, many studies have incorporated exteroceptive sensors such as LiDAR and depth cameras, enabling more sophisticated terrain traversal~\cite{hl:wang2025more, hl:long2025learning, allshire2025visual} and object interaction~\cite{physhsi, hdmi, zhang2025falcon}.
Our work further investigates the task of humanoid skateboarding, which demands significantly higher levels of control agility and whole-body coordination.

\subsection{Legged Robots on Dynamic Platforms}

The interaction between legged robots and dynamic platforms poses significant challenges. Model-based approaches have enabled quadrupedal robots to perform skateboarding through offline trajectory optimization~\cite{xu2024optimization}. Learning-based methods further allow quadrupeds to acquire more robust skateboarding skills, including mounting~\cite{belov2025quadrupedal} and maneuvering the board~\cite{liu2025discrete}. By simulating dynamic platform motion in virtual environments, \citet{huang2025learning} demonstrated effective balance control across a variety of unstable surfaces.

Humanoid skateboarding, however, remains especially demanding for model-based frameworks~\cite{takasugi2018extended, takasugi2016real,mikkola2024online}, as it requires agile, real-time control and highly accurate physical modeling~\cite{varszegi2015skateboard, kuleshov2006mathematical}. In contrast, while RL has shown great potential for enabling robust and dexterous humanoid behaviors in other domains~\cite{an2026ai,gu2026humanoid}, its application to skateboarding remains largely unexplored. Our framework leverages these capabilities to realize learning-based agile skateboarding skills on a humanoid robot in the real world.
\section{Humanoid-Skateboard System}

\subsection{System Modeling}

\begin{figure}[t]
    \centering
    \includegraphics[width=1.0\linewidth]{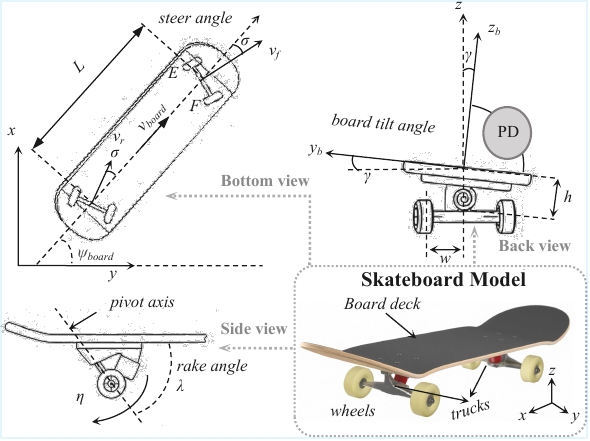}
    \caption{\textbf{Skateboard Model.} 
We analyze the skateboard kinematic structure and derive the coupling relationships among the board tilt, truck steering, and rake angles, which form the basis of the lean-to-steer behavior. }
    \label{fig:model}
\end{figure}

We consider a humanoid robot interacting with a fully passive articulated skateboard via foot contacts, with all actuation provided by the humanoid.
As illustrated in Fig.~\ref{fig:model}, the skateboard comprises three main components: the deck, trucks, and wheels. The deck supports the humanoid’s weight and transmits forces to the trucks, which connect the deck to the wheels. In real skateboards, the trucks use a kingpin-based suspension with elastic bushings, which converts deck tilting into truck steering while providing force damping.  
While directly modeling the full suspension in simulation is challenging due to its structural complexity and compliance. Therefore, we adopt a simplified kinematic model that preserves the key lean-to-steer behavior while remaining tractable for control and simulation.

Under this formulation, wheel–ground contacts are modeled as non-holonomic rolling constraints that restrict lateral slip. Instead of explicitly resolving the constraint forces, we exploit the truck geometry to relate the board tilt angle $\gamma$ to the rotation of the truck axes. Due to the mechanical structure of the trucks, tilting the board by $\gamma$ induces a coordinated rotation of the truck axes, resulting in a kinematic coupling \cite{varszegi2015skateboard} described by:
\begin{equation}
\label{eq:angle}
\tan \sigma = \tan \lambda \, \sin \gamma,
\end{equation}
where $\lambda$ is the constant rake angle of the skateboard, and $\sigma$ is the resulting truck steering angle. Intuitively, this implies that the truck steering angle is determined by the board tilt angle, with larger tilts producing greater steering deflections. Detailed derivation is provided in Appendix A.

This abstraction directly links board tilt to truck steering, capturing the essential equality constraint of humanoid-driven maneuvers while avoiding unnecessary mechanical redundancies. The resulting model is both computationally efficient and physically representative, providing a tractable foundation for subsequent control and learning of humanoid skateboarding.

\subsection{Humanoid Skateboarding Task}
\begin{figure*}
    \centering
    \includegraphics[width=\linewidth]{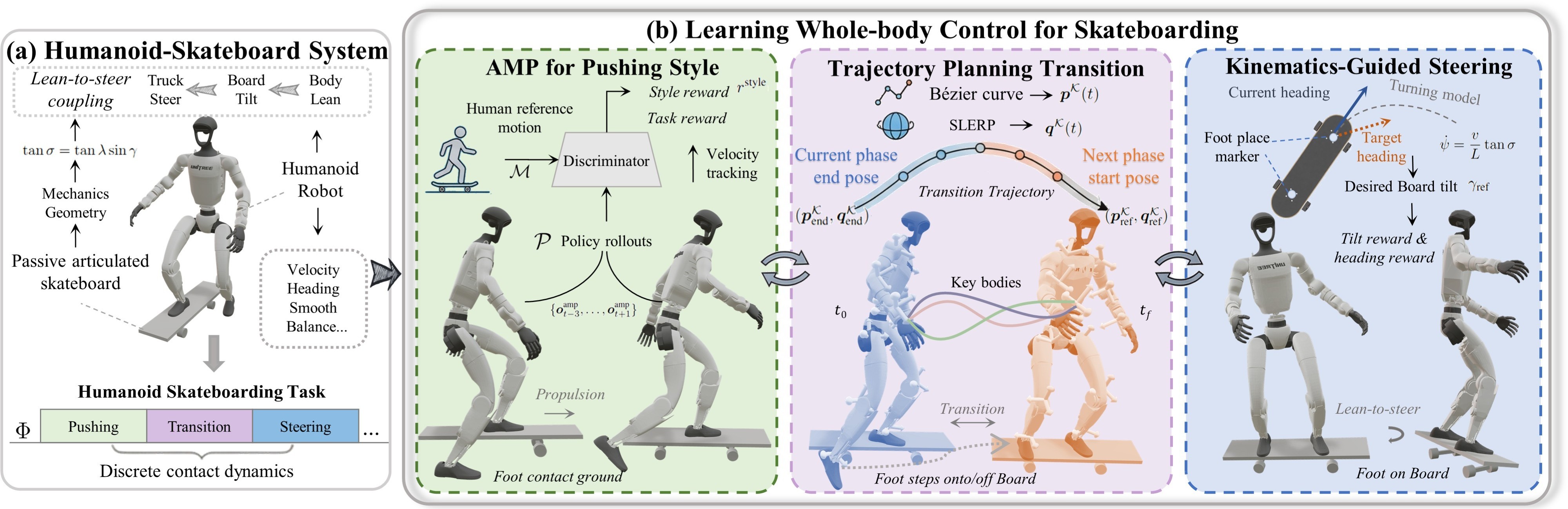}
    \caption{\textbf{Framework of \ours{}.} 
(a) We first analyze and model the humanoid–skateboard system, deriving a physics-inspired lean-to-steer coupling mechanism.
Due to the distinct contact dynamics and control objectives across skateboarding phases, we adopt a phase-wise learning strategy.
(b) The learning-based whole-body control framework integrates an AMP-based pushing style for active forward propulsion, a steering strategy guided by kinematics-aware tilt references, and a trajectory-guided transition mechanism to enable stable switching between pushing and steering phases.}
    \label{fig:framework}
\end{figure*}

We formulate humanoid skateboarding as a hybrid control task, comprising two periodically recurring phases, \textit{pushing} and \textit{steering}, which arise from discrete changes in the system’s contact topology. Each phase is governed by distinct dynamics and control objectives. The framework is shown in Fig.~\ref{fig:framework}.

During the pushing phase (Section~\ref{sec:push}), one foot of the humanoid maintains contact with the skateboard to stabilize the body, while the other intermittently contacts the ground to generate propulsion. Forward motion is produced by tangential ground reaction forces at the pushing foot, whereas the supporting foot ensures balance on the moving board.

In the steering phase (Section~\ref{sec:steer}), both feet of the humanoid remain on the skateboard, producing a passive gliding motion. Steering is achieved by modulating the board tilt through body leaning, which induces truck rotation and allows the humanoid to follow a desired heading.

While transitions between pushing and steering are challenging due to the distinct dynamics and control objectives of each phase. In Section~\ref{sec:trans}, we describe strategies to explicitly bridge these phases, enabling smooth and stable phase transitions during humanoid skateboarding.

\section{Method}

\subsection{Problem Formulation}

In this work, we adopt the Unitree G1 humanoid robot~\cite{unitree-g1} with 23 controllable degrees of freedom (DoF), excluding the three DoFs of each wrist. We formulate the humanoid skateboarding task as a RL problem, in which the agent interacts with the environment through a policy $\pi$ to maximize cumulative reward. At each timestep $t$, the policy receives the state $\bm s_t$ and outputs an action $\bm a_t \in \mathbb{R}^{23}$, which is subsequently mapped to motor torques via a PD controller. This defines the policy as $\pi(\bm a_t | \bm s_t)$. The environment dynamics $p(\bm s_{t+1} | \bm s_t, \bm a_t)$ determine the next state, while a dense reward $r( \bm s_t, \bm a_t)$ evaluates skateboarding performance while providing regularization. The agent aims to maximize the expected return $J = \mathbb{E}\Big[\sum_{t=0}^{T-1} \beta^t r_t \Big]$, where $\beta$ denotes the discount factor.

We utilize an asymmetric actor–critic framework for policy training, where the actor observes only proprioceptive information, while the critic has access to additional privileged information.
At each timestep $t$, the policy state $\bm s_t$ consists of a short history of proprioceptive states 
$\bm o^{\text{prop}}_{t-4:t}$, where each proprioceptive state 
$\bm o^{\text{prop}}_{t} \in \mathbb{R}^{78}$ is defined as
\begin{equation}
\bm o^{\text{prop}}_t =
[
\bm{c}_t, \bm{\omega}_{t}, \bm{g}_{t}, \bm{\theta}_t, \dot{\bm{\theta}}_t, \bm{a}_{t-1}, \Phi
],
\end{equation}
where $\bm{c}_t=[v_{cmd}, \psi] \in \mathbb{R}^{2}$ contains the desired skateboard forward velocity $ v_{cmd}$ and heading target $\psi$, $\bm{\omega}_t \in \mathbb{R}^{3}$ is the base angular velocity, $\bm{g}_t \in \mathbb{R}^{3}$ is the base projected gravity, $\bm{\theta}_t \in \mathbb{R}^{23}$ and $\dot{\bm{\theta}}_t \in \mathbb{R}^{23}$ are the joint angles and velocities, and $\bm{a}_{t-1} \in \mathbb{R}^{23}$ is the previous action. The phase cycle is represented by a normalized variable $\Phi= (t \bmod H) / H \in [0,1)$, where $H$ is the cycle duration. This periodic variable governs transitions between pushing and steering phases, serving as a temporal reference for the policy.

To expose the skateboard state to the critic during training, we define the privileged observation $\bm o^{\text{priv}}_{t}$ as
\begin{equation}
\bm o^{\text{priv}}_t =
[
\bm{v}_t, \bm{p}^{b}_t, \bm{r}^{b}_t, \bm{v}^{b}_t, \bm{\omega}^{b}_t,\bm{\theta}^{{b}}_t,\bm{f}^{g}_t,\bm{f}^{b}_t
],
\end{equation}
where $\bm{v}_t \in \mathbb{R}^{3}$ denotes the robot base linear velocity, $\bm{p}^{b}_t\in \mathbb{R}^{3}$, $\bm{r}^{b}_t\in \mathbb{R}^{6}$, $\bm{v}^{b}_t\in \mathbb{R}^{3}$ and $\bm{\omega}^{b}_t\in \mathbb{R}^{3}$ represent the skateboard base position, orientation, linear velocity and angular velocity in the robot's frame, respectively. $\bm{\theta}^{{b}}_t\in \mathbb{R}^{3}$ is the skateboard's tilt and truck joint angles, 
while $\bm{f}^{{g}}_t\in \mathbb{R}^{6}$ and $\bm{f}^{{b}}_t\in \mathbb{R}^{6}$ represent the humanoid's feet contact forces with ground and skateboard, respectively.
To guide the humanoid toward the desired behaviors in each phase, we detail phase-specific reward functions in Appendix C. The overall reward is formulated as:
\begin{equation}
r_t = \mathbb{I}^{\text{push}} \cdot r_t^{\text{push}}
     + \mathbb{I}^{\text{steer}} \cdot r_t^{\text{steer}}
     + \mathbb{I}^{\text{trans}} \cdot r_t^{\text{trans}}
     + r_t^{\text{reg}},
\end{equation}
where $\mathbb{I}^{\text{push}}$, $\mathbb{I}^{\text{steer}}$, and $\mathbb{I}^{\text{trans}}$ are binary indicators denoting the current contact phase, and $r_t^{\text{reg}}$ is a global regularization term.

\subsection{Adversarial Motion Prior for Pushing Style}
\label{sec:push}
During the pushing phase, the primary objective is to achieve a target skateboard velocity $ v_{cmd}$ by generating physically plausible propulsion through intermittent foot-ground contacts. Traditional reference-tracking methods enforce strict adherence to predefined joint trajectories \cite{peng2018deepmimic}, resulting in rigid behaviors that limit adaptation to varying environments and velocity demands. To encourage more natural and robust behaviors, we build on the AMP framework, which employs a discriminator $D_\phi$ to distinguish policy rollouts from reference human pushing motions.

The AMP observation is defined as the robot joint angles
$\bm{o}_t^{\text{amp}} = \theta_t\in \mathbb{R}^{23}$.
To provide the discriminator with temporal context, we construct a motion window
$\tau_t = \{\bm{o}_{t-3}^{\text{amp}}, \ldots, \bm{o}_{t+1}^{\text{amp}}\}$ spanning five timesteps. Reference transitions are sampled from a human motion dataset $\mathcal{M}$, while generated transitions are collected from the current policy rollouts $\mathcal{P}$.

The discriminator is trained to distinguish between expert and agent motions using a least-squares loss, augmented with a gradient penalty to promote training stability:
\begin{align}
\label{eq:disc_loss}
\arg\max_{\phi} \ \ & 
\mathbb{E}_{\tau \sim \mathcal{M}}[(D_{\phi}(\tau) - 1)^2] 
+ \mathbb{E}_{\tau \sim \mathcal{P}}[(D_{\phi}(\tau) + 1)^2] \nonumber \\
& + \frac{\alpha^{d}}{2} \mathbb{E}_{\tau \sim \mathcal{M}}[\|\nabla_{\phi} D_{\phi}(\tau)\|_2].
\end{align}

The discriminator output $d = D_{\phi}(\tau_t)$ is then mapped to a bounded style reward:
\begin{equation}
r^{\text{style}}(s_t) = \alpha \cdot \max\left(0,\ 1 - \frac{1}{4}(d - 1)^2\right),
\end{equation}
where $\alpha$ is a scaling coefficient.
The total reward for the pushing phase combines task-specific and style components:
\begin{equation}
r_t^{\text{push}} = r_t^{\text{task}} + r_t^{\text{style}}.
\end{equation}

Here, $r_t^{\text{task}}$ encourages accurate tracking of the commanded skateboard velocity, while $r_t^{\text{style}}$ guides the humanoid to exhibit human-like pushing behaviors.

\subsection{Kinematics-Guided Heading-Oriented Steering}
\label{sec:steer}

In the steering phase, the humanoid receives a target heading command $\psi$ and leans its body to induce a board tilt $\gamma$, which, through the truck geometry coupling in Eq.~\eqref{eq:angle}, produces the corresponding truck steering angle $\sigma$ and achieves the desired turning behavior. To exploit this physical coupling, we provide the humanoid with a kinematics-guided reference specifying the desired board tilt angle for heading regulation.

Assuming a planar kinematic model of the skateboard, we adopt a bicycle-model approximation commonly used in vehicle dynamics~\cite{yang2013overview}, which captures the yaw motion of the skateboard while neglecting lateral slip and vertical dynamics. Under this formulation, the yaw rate $\dot{\psi}$ is given by:

\begin{equation}
\dot{\psi} = \frac{v}{L} \tan \sigma,
\label{eq:yaw_rate}
\end{equation}
where $v$ is the forward velocity of the skateboard in its base frame, and $L$ is the wheelbase. Combining this with the lean-to-steer relationship in Eq.~\eqref{eq:angle}, we obtain

\begin{equation}
\dot{\psi} = \frac{v}{L} \tan \lambda \, \sin \gamma,
\label{eq:yaw_rate_gamma}
\end{equation}
which links the board tilt angle $\gamma$ to the resulting yaw rate.

Given a desired heading change $\Delta\psi = \psi - \psi_{\text{board}}$ over a steering horizon $\Delta t$, where $\psi_{\text{board}}$ denotes the current heading of the skateboard, we assume a constant yaw rate $\dot{\psi} \approx \Delta\psi / \Delta t$ to achieve a smooth and gradual turn. Substituting this into Eq.~\eqref{eq:yaw_rate_gamma} yields the kinematics-guided tilt reference:

\begin{equation}
\gamma_{\text{ref}} = \arcsin\!\left( \frac{L \, \Delta\psi}{v \, \Delta t \, \tan \lambda} \right),
\label{eq:gamma_ref}
\end{equation}
where $\gamma_{\text{ref}}$ represents the desired board tilt angle that minimizes the heading error. To ensure numerical stability at low speeds or large desired turns, we clip $v$ to a minimum threshold and constrain $\gamma_{\text{ref}}$ within feasible lean limits.

We define a tilt reward to guide the policy toward this physically consistent lean:
\begin{equation}
r_{\text{tilt}} = \exp\!\left( - \frac{\|\gamma - \gamma_{\text{ref}}\|^2}{\sigma_\gamma^2} \right),
\label{eq:tilt_reward}
\end{equation}
where $\sigma_\gamma$ sets the tolerance. This is combined with a heading-tracking reward
$r_{\text{heading}} = \exp( - (\Delta \psi)^2 / \sigma_\psi^2 )$,
ensuring that the policy aligns the skateboard with the target direction while maintaining a physically feasible lean angle.

To further promote balance and stable foot placement during steering, two virtual markers are defined above the skateboard trucks. These markers indicate preferred feet contact locations, and the policy is encouraged to place the feet close to them, providing auxiliary guidance for the humanoid’s stance.

\subsection{Trajectory Planning for Phase Transition}
\label{sec:trans}

Direct transitions between the pushing and steering phases are challenging due to differences in body poses, contact dynamics, and learning objectives. Policies trained without explicit transition guidance often fail to explore new contact phases, which can lead to convergence to local optima~\cite{zhang2024wococolearningwholebodyhumanoid}.

To address this problem, we introduce a \emph{trajectory-guided transition} mechanism that explicitly bridges the pushing and steering phases by generating intermediate reference states. Specifically, we select a set of key bodies, denoted by $\mathcal{K}$, which capture the humanoid's essential motion during phase transitions. The positions and orientations of these key bodies relative to the skateboard are then used to plan smooth and physically consistent transition trajectories.

At the end of the current phase, the humanoid’s key body poses $(\bm{p}^{\mathcal{K}}_{\text{end}}, \bm{q}^{\mathcal{K}}_{\text{end}})$ are obtained online and serve as the initial condition for trajectory planning. Unlike predefined motions, these terminal poses vary across episodes and policies, resulting in a state-dependent transition problem.

For each phase, we also define a canonical stable pose 
$(\bm{p}^{\mathcal{K}}_{\text{ref}}, \bm{q}^{\mathcal{K}}_{\text{ref}})$ 
relative to the skateboard frame, representing the nominal start configuration of the subsequent phase. The transition is therefore formulated as a trajectory connecting the online terminal pose of the current phase to the fixed reference pose of the next phase. For example, the end pose of the pushing phase is planned toward the steering phase reference pose, and vice versa.

For body translations, the planned Cartesian position at timestep $t$ is generated using an $n$-th order Bézier curve:

\begin{equation}
\bm{p}^{\mathcal{K}}(t) = \sum_{i=0}^{n} \binom{n}{i} (1 - s)^{\,n-i} s^i \, \bm{p}^{\mathcal{K}}_i, 
\quad s = \frac{t - t_0}{t_f - t_0},
\end{equation}
where $\bm{p}^{\mathcal{K}}_0 = \bm{p}^{\mathcal{K}}_{\text{end}}$ and 
$\bm{p}^{\mathcal{K}}_n = \bm{p}^{\mathcal{K}}_{\text{ref}}$ denote the initial and target positions, and $\{\bm{p}^{\mathcal{K}}_i\}_{i=1}^{n-1}$ are intermediate control points shaping the trajectory. Here, $t_0$ and $t_f$ are the start and end times of the transition.

Boy orientations are interpolated using spherical linear interpolation (slerp) between quaternions:
\begin{equation}
\bm{q}^{\mathcal{K}}(t) =
\frac{\sin((1-s)\Omega)}{\sin \Omega} \, \bm{q}^{\mathcal{K}}_{\text{end}} +
\frac{\sin(s \, \Omega)}{\sin \Omega} \, \bm{q}^{\mathcal{K}}_{\text{ref}},
\end{equation}
where $s = \frac{t - t_0}{t_f - t_0}$ and 
$\Omega = \arccos(\langle \bm{q}^{\mathcal{K}}_{\text{end}}, 
\bm{q}^{\mathcal{K}}_{\text{ref}} \rangle)$ is the angular distance between the quaternions. This ensures smooth and physically consistent orientation transitions while stepping onto or off the skateboard.

At each transitional timestep $t$, the policy receives a tracking reward that encourages it to follow the planned trajectories $(\bm{p}^{\mathcal{K}}(t),\bm{q}^{\mathcal{K}}(t))$. This trajectory-guided guidance shapes the exploration space, enabling reliable phase switching while retaining sufficient flexibility for the policy to adapt to dynamic interactions. By integrating trajectory-guided transitions with phase-specific learning objectives, the humanoid achieves stable and repeatable transitions between pushing and steering, facilitating consecutive skateboarding behaviors.

\subsection{Sim-to-Real Transfer}
\subsubsection{Skateboard Physical Identification}

To bridge the sim-to-real gap, we identify an equivalent PD model for the skateboard’s passive tilt dynamics (back view of Fig.~\ref{fig:model}) through a sequential analytical identification procedure based on its free-decay roll response~\cite{modern_control_engineering}. Specifically, the skateboard is perturbed in roll and released, allowing the truck bushings and pivot friction to produce a naturally decaying oscillation.
We first quantify energy dissipation by measuring two successive roll angle peaks, $\phi(t)$ and $\phi(t+T)$. The logarithmic decrement $\delta$ and the corresponding damping ratio $\zeta$ are computed as
\begin{equation}
\delta = \ln\frac{\phi(t)}{\phi(t+T)} , \quad
\zeta = \frac{\delta}{\sqrt{4\pi^2 + \delta^2}} ,
\end{equation}
where $T$ is the observed oscillation period. This damping ratio captures the intrinsic dissipative effects of the truck bushings and pivot interfaces.
Given the identified $\zeta$, we estimate the equivalent torsional stiffness of the truck assembly by matching the observed roll oscillation frequency. The undamped natural frequency is computed as:
\begin{equation}
\omega_n = \omega_d / \sqrt{1 - \zeta^2},
\end{equation}
where $\omega_d = 2\pi/T$ is the damped frequency. Approximating the skateboard’s roll inertia $I$ using a rigid cuboid model dominated by the deck geometry, the torsional stiffness is identified as $k = I \omega_n^2$, which represents the restorative torque generated by the bushings under roll deformation.
Finally, the velocity-dependent damping coefficient is obtained as $d = 2 \zeta \sqrt{k I}$, modeling the dissipative torque arising from bushing hysteresis and pivot friction. Embedding the identified parameters $(k, d)$ into simulation yields a physically grounded tilt spring–damper model that faithfully reproduces the passive truck mechanics observed on the real skateboard.

\subsubsection{Domain Randomization}
To further improve policy robustness and facilitate effective sim-to-real transfer, we employ domain randomization (DR) during training. The detailed configurations are provided in Table~\ref{tab:dr}.  

\begin{table}
\centering
\caption{Domain Randomization Parameters.}
\begin{tabular}{llc}
\toprule
\textbf{DR terms} & \textbf{Range} & Unit \\
\midrule
Robot Center of Mass & $\mathcal{U}(-2.5,\ 2.5)$ & $cm$ \\
Skateboard Center of Mass & $\mathcal{U}(-2.5,\ 2.5)$ & $cm$ \\
Default Root Position & $\mathcal{U}(-2.0,\ 2.0)$ & $cm$ \\
Default Joint Position & $\mathcal{U}(-0.01,\ 0.01)$ & $rad$ \\
Push Robot Base & $\mathcal{U}(-0.5,\ 0.5)$  & $m/s$ \\
Robot Body Friction & $\mathcal{U}(0.3,\ 1.6)$ & - \\
Skateboard Deck Friction & $\mathcal{U}(0.8,\ 2.0)$ & - \\

\bottomrule
\end{tabular}
\label{tab:dr}
\end{table}
\section{Experiments}

\begin{table*}[htbp]
  \centering
  \caption{Simulation Results.}
  \label{tab:Simulation Results}
  \resizebox{0.8\textwidth}{!}{
  \begin{tabular}{lccccccc}
    \toprule
    {\textbf{Method}} 
     & $E_{\text{succ}}\!\uparrow$ & $E_{\text{vel}}\!\downarrow$ & $E_{\text{yaw}}\!\downarrow$ &  $E_{\text{smth}}\!\downarrow$  &  $E_{\text{contact}}\!\downarrow$  \\
    \midrule
    \rowcolor{gray!18}\multicolumn{7}{l}{\textbf{Ablation on Pushing Style}} \\
    \ours{}-Tracking-Based  & $11.12\!\pm\! \mathsmaller{3.86}$ & $0.435\!\pm\! \mathsmaller{0.101}$ & ${0.568}\!\pm\! \mathsmaller{0.092}$ & ${0.044}\!\pm\! \mathsmaller{0.025}$ &  ${0.015}\!\pm\! \mathsmaller{0.010}$ \\
    \ours{}-Gait-Based  & $82.38\!\pm\! \mathsmaller{7.25}$ & $0.102\!\pm\! \mathsmaller{0.035}$ & ${0.302}\!\pm\! \mathsmaller{0.041}$ & ${0.043}\!\pm\! \mathsmaller{0.011}$ &  ${0.130}\!\pm\! \mathsmaller{0.072}$ \\
    \ours{} (ours) & $\textbf{{100.00}}\!\pm\! \mathsmaller{0.00}$ & $\textbf{{0.056}}\!\pm\! \mathsmaller{0.013}$ & $\textbf{{0.208}}\!\pm\! \mathsmaller{0.014}$ & $\textbf{0.033}\!\pm\! \mathsmaller{0.005}$ &  $\textbf{0.001}\!\pm\! \mathsmaller{0.001}$ \\
    \midrule
    \rowcolor{gray!18}\multicolumn{7}{l}{\textbf{Ablation on Steering Strategy}} \\
    \ours{}-w/o-Tilt Guidance   & $96.72\!\pm\! \mathsmaller{2.10}$ & $0.071\!\pm\! \mathsmaller{0.010}$ & ${0.233}\!\pm\! \mathsmaller{0.027}$ & $0.035\!\pm\! \mathsmaller{0.017}$ &  $0.002\!\pm\! \mathsmaller{0.002}$\\
    \ours{} (ours) & $\textbf{{100.00}}\!\pm\! \mathsmaller{0.00}$ & $\textbf{{0.056}}\!\pm\! \mathsmaller{0.013}$ & $\textbf{{0.208}}\!\pm\! \mathsmaller{0.014}$ & $\textbf{0.033}\!\pm\! \mathsmaller{0.005}$ &  $\textbf{0.001}\!\pm\! \mathsmaller{0.001}$ \\
    \midrule
    \rowcolor{gray!18}\multicolumn{7}{l}{\textbf{Ablation on Transition Mechanism}} \\
    \ours{}-AMP Transition & $85.12\!\pm\! \mathsmaller{4.11}$ & $\textbf{0.053}\!\pm\! \mathsmaller{0.025}$ & $ {0.265}\!\pm\! \mathsmaller{0.050}$ &$0.040\!\pm\! \mathsmaller{0.007}$ &  $0.394\!\pm\! \mathsmaller{0.015}$ & \\
    \ours{}-Translation-only & ${89.55}\!\pm\! \mathsmaller{2.30}$ & $0.064\!\pm\! \mathsmaller{0.020}$ & ${0.294}\!\pm\! \mathsmaller{0.075}$ &$0.039\!\pm\! \mathsmaller{0.012}$ &  $0.038\!\pm\! \mathsmaller{0.012}$ & \\
    \ours{}-Mixed Initialization& ${86.03}\!\pm\! \mathsmaller{1.70}$ & $0.067\!\pm\! \mathsmaller{0.024}$ & ${0.278}\!\pm\! \mathsmaller{0.092}$ &$0.037\!\pm\! \mathsmaller{0.015}$ &  $0.371\!\pm\! \mathsmaller{0.025}$ & \\
    \ours{} (ours) & $\textbf{{100.00}}\!\pm\! \mathsmaller{0.00}$ & $0.056\!\pm\! \mathsmaller{0.013}$ & $\textbf{{0.208}}\!\pm\! \mathsmaller{0.014}$ & $\textbf{0.033}\!\pm\! \mathsmaller{0.005}$ &  $\textbf{0.001}\!\pm\! \mathsmaller{0.001}$ \\
    \bottomrule
  \end{tabular}}
\end{table*}

\subsection{Experiment Setup}

In simulation, we use a standard $80 cm \times 20 cm \times  12 cm$ skateboard model, the coupling mechanism in Eq.~\eqref{eq:angle} is embedded into the simulator as an equivalent constraint. The details of skateboard model are provided in Appendix B. All trainings and evaluations are implemented in mjlab~\cite{zakka2026mjlablightweightframeworkgpuaccelerated}, which integrates MuJoCo physics~\cite{todorov2012mujoco} with the IsaacLab API~\cite{mittal2025isaac}. Humanoid skateboarding policies are trained using the PPO~\cite{ppo} algorithm across 4,096 parallel environments, with each episode lasting 20 seconds. For each experiment, we evaluate 1,000 rollout episodes in simulation and report results over five random seeds. Additional training details are given in Appendix D. In real-world experiments, the policy is deployed on a Unitree G1 humanoid robot at 50 Hz, with joint positions tracked by a 500 Hz PD controller.

\subsubsection{Evaluation Metrics}
We adopt the following metrics to evaluate the humanoid skateboarding performance:

\begin{itemize}
    \item \textbf{Task success rate} ($E_{\text{succ}}$): Defined as the percentage of episodes completed without termination.
    
    \item \textbf{Velocity tracking error} ($E_{\text{vel}}$): Measured as the mean absolute error between the commanded velocity $ v_{\text{cmd}}$ and the actual skateboard forward velocity $ v_{\text{board}}$ during the pushing phase:
    $E_{\text{vel}} = \mathbb{E} \left[ \left| v_{\text{cmd}} - v_{\text{board}} \right| \right]$.
    
    \item \textbf{Heading tracking error} ($E_{\text{yaw}}$): Quantifies the accuracy of heading control during steering phase using the heading error:
    $E_{\text{yaw}} = \mathbb{E} \left[\left| \psi - \psi_{\text{board}} \right| \right]$.
    
    \item \textbf{Motion smoothness} ($E_{\text{smth}}$): Evaluated by aggregating joint angle variations across consecutive control steps, promoting temporally smooth and plausible behaviors.
    
    \item \textbf{Contact error} ($E_{\text{contact}}$): Defined as the per-step violation of the foot–board contact pattern, i.e., single-foot contact during pushing and double-foot contact during steering.
\end{itemize}

\subsubsection{Baselines} 
To evaluate the effectiveness of \ours{}, we compare it with the following ablated baseline variants:

\begin{itemize}
    \item \textbf{Pushing style}: We consider two baselines: (i) a \textbf{\emph{Tracking-based}} variant~\cite{peng2018deepmimic} that directly tracks reference skateboarding motions from $\mathcal{M}$, and (ii) a \textbf{\emph{Gait-based}} variant~\cite{TienKungLab2024} in which the humanoid’s pushing follows a fixed gait schedule.

    \item \textbf{Steering strategy}: We compare our steering design with a \textbf{\emph{w/o Tilt Guidance}} baseline, where the policy implicitly learn board tilt angles for steering without physics-guided tilt reference.

    \item \textbf{Transition mechanism}: We evaluate three variants: (i) \textbf{\emph{AMP Transition}}, which uses phase-specific reference motions, including stepping onto and off the skateboard, and relies on style rewards to guide transitions, (ii) \textbf{\emph{Translation-only}}, which guides only translational motion while ignoring orientation, underscoring the importance of jointly controlling both position and rotation during phase transitions, and (iii) \textbf{\emph{Mixed Initialization}}~\cite{liu2025discrete} utilizes a balanced 50/50 mixture of pushing and steering phases during reset to facilitate comprehensive phase exploration.

\end{itemize}

\subsection{Main Results}

Table~\ref{tab:Simulation Results} summarizes the overall performance of \ours{} compared to the baselines across metrics. These results demonstrate that \ours{} enables robust humanoid skateboarding with stable, accurate, smooth motions and continuous, reliable phase transitions. The contributions of the key design choices are summarized as follows:

\textbf{AMP enables flexible and natural pushing.} 
For pushing, the \emph{Tracking-based} baseline strictly follows reference motions, limiting its adaptability to varying contact timings and velocities and leading to the lowest success rate. The \emph{Gait-based} baseline, lacking human motion priors, produces less smooth motions and higher contact errors, reflecting its difficulty in generating coordinated, human-like pushing behaviors.

\textbf{Physics-guided steering enhances heading control.} 
For steering, the \emph{w/o Tilt Guidance} baseline learns board tilt angles implicitly, which reduces heading accuracy and highlights the importance of explicitly incorporating the lean-to-steer coupling for precise directional control.

\textbf{Trajectory guidance is essential for phase transitions.} 
For phase transitions, the \emph{AMP Transition} baseline, which relies solely on full reference motions, achieves moderate success rates but incurs significant contact errors, as the robot fails to explore proper phase switching. The \emph{Translation-only} baseline, which guides translation while omitting orientation alignment, exhibits poor heading tracking, causing the robot to align with the board direction rather than adopt a side-on posture and making steering control difficult.

\subsection{More Analysis}

\begin{figure}[t]
    \centering
    \includegraphics[width=\linewidth]{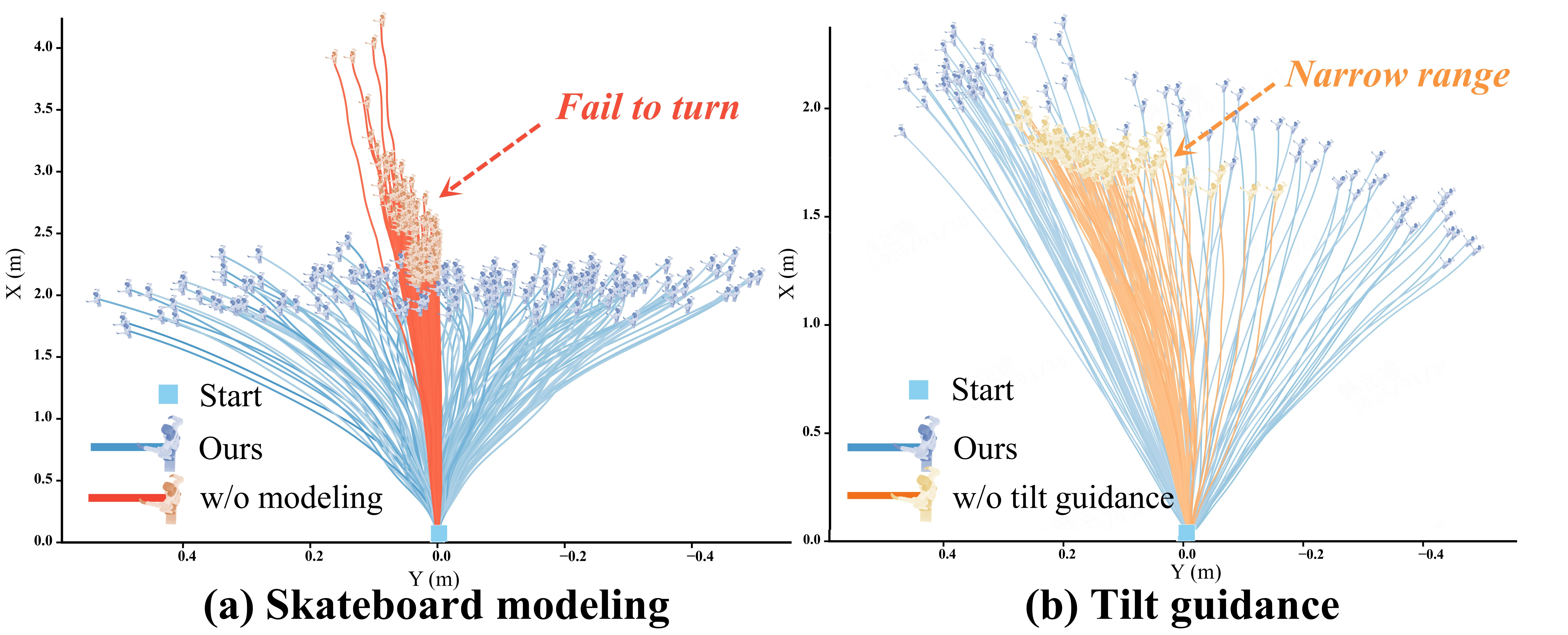}
    \caption{\textbf{Steering Trajectories Visualizations.} 
(a) Omitting lean-to-steer coupling prevents effective steering. 
(b) Incorporating physics-guided tilt guidance substantially increases the reachable heading range and precision.
}
    \label{fig:tilt}
\end{figure}

\textbf{Skateboard Modeling (Fig.~\ref{fig:tilt}).}
We validate the effect of skateboard modeling and visualize the resulting steering trajectories in Fig.~\ref{fig:tilt} (a). Similar to prior simplified models~\cite{liu2025discrete}, omitting the equality constraint in Eq.~\eqref{eq:angle} prevents board tilting from inducing truck steering, leaving the skateboard able only to glide straight forward with negligible turning capability.

Fig.~\ref{fig:tilt} (b) compares policies trained without tilt guidance under identical steering commands by measuring the range of reachable headings. Without tilt guidance, the achievable heading range is narrow. In contrast, incorporating tilt guidance produces smooth turning trajectories and enables the humanoid to reach a substantially wider range of headings with higher precision. These results demonstrate that tilt guidance is essential for effective and precise turning control.

\begin{figure}[t]
    \centering
    \includegraphics[width=\linewidth]{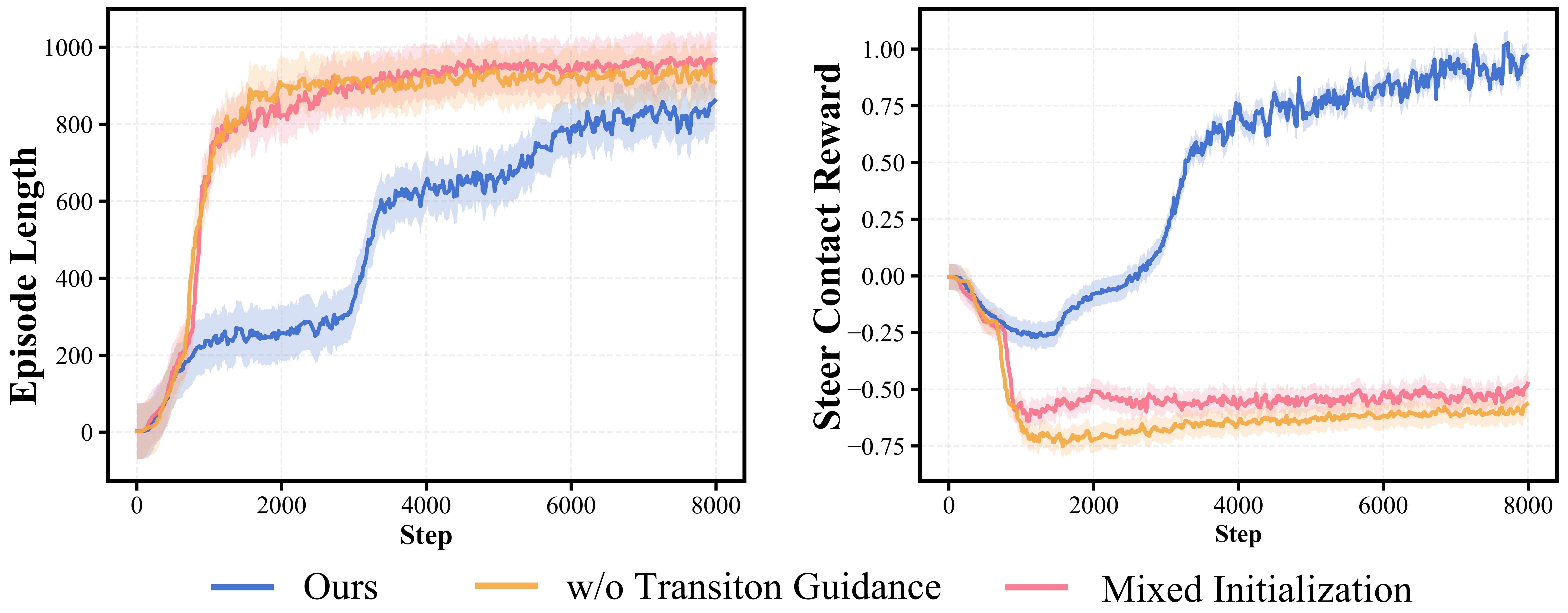}
    \caption{\textbf{Training Performance Comparison.} 
Episode length (left) and steering contact reward (right). Without transition guidance, the policy fails to maintain correct foot–board contacts. In contrast, the policy discovers correct contact patterns early in training and establishes stable phase transitions.}
    \label{fig:reward}
\end{figure}

\textbf{Phase Exploration (Fig.~\ref{fig:reward}).}
To further analyze the necessity of transition guidance, we train policies without it under two settings: (i) all episodes are initialized to start in the pushing phase, matching our setup, and (ii) a 50/50 mixture of pushing and steering phases, as used in prior work~\cite{liu2025discrete} to promote phase exploration. We measure episode length and a steering contact reward, which encourages double-foot board contact and penalizes ground contact during steering.

In both settings, episode length increases rapidly during early training, yet the steering contact reward remains low, indicating persistent incorrect foot–board contact patterns. This suggests that policies trained without transition guidance fail to learn phase transitions, collapsing to a trivial pushing-only behavior and failing to properly execute the steering phase, even under mixed initialization. In contrast, \ours{} discovers correct contact patterns by mid-training, successfully learns foot-mounting transitions, and achieves higher rewards, demonstrating that trajectory-guided transitions are essential for enabling phase switching and avoiding local optima.

\begin{figure}[t]
    \centering
    \includegraphics[width=\linewidth]{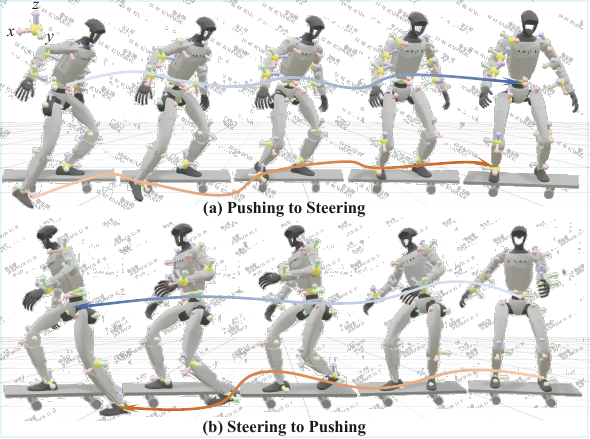}
    \caption{\textbf{Transition trajectories analysis.} 
Representative trajectories during phase transitions. The humanoid maintains smooth, coordinated whole-body motions, with seamless transitions between pushing and steering. }
    \label{fig:traj}
\end{figure}

\textbf{Trajectories Analysis (Fig.~\ref{fig:traj}).}
We further examine the humanoid's skateboarding transitions by analyzing collected trajectory sequences. As shown in Fig.~\ref{fig:traj}, the robot maintains smooth, coordinated whole-body motions with seamless transitions between pushing and steering. The trajectories display consistent foot placement and gradual body pose adjustments, reflecting strong temporal coherence and physical plausibility enabled by our trajectory-guided transitions.

\subsection{Real-World Experiments}

\subsubsection{Overall Performance}

Overall, Fig.~\ref{fig:cover} and the supplementary videos demonstrate that \ours{} achieves robust real-world skateboarding across diverse conditions. The system executes complete skateboarding behaviors including pushing, steering, and phase transitions, generalizes to different skateboard platforms, and performs reliably in both indoor and outdoor environments. Lean-to-steer control emerges naturally through body tilting, enabling heading regulation. The humanoid maintains smooth and coordinated foot placements during transitions, withstands external disturbances, and sustains multiple continuous skateboarding cycles.

\subsubsection{Transition Details}

\begin{figure}[t]
    \centering
    \includegraphics[width=1.0\linewidth]{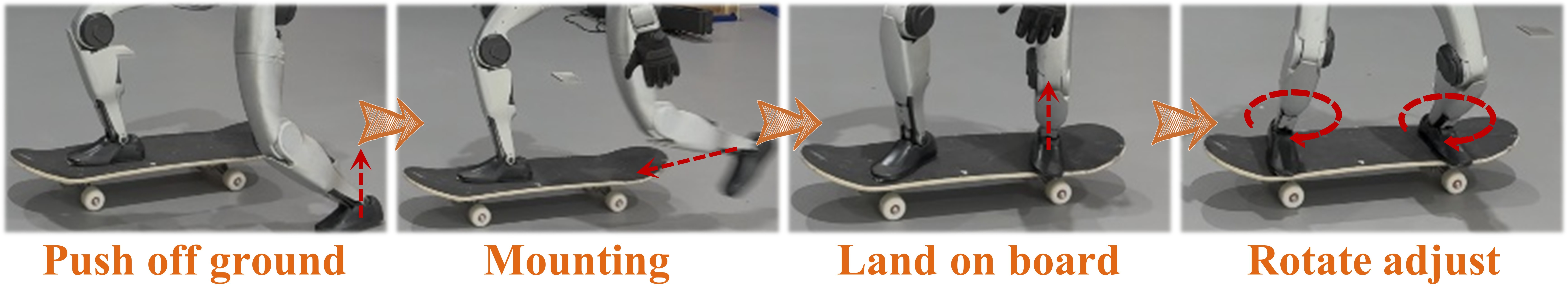}
    \caption{\textbf{Details of feet motions during transitions.} The humanoid pushes off the ground, mounts the skateboard, and adjusts orientation on the board.}
    \label{fig:detail}
\end{figure}

We analyze the transitions from pushing to steering, focusing on detailed humanoid feet motion. In Fig.~\ref{fig:detail}, the transition begins with the foot pushing against the ground to generate propulsion, followed by lifting and placing the foot onto the skateboard. Once on the board, the humanoid performs in-place adjustments, rotating the body to align the torso perpendicular to the skateboard deck, thereby facilitating stable steering. These motions closely match human behaviors and are consistent with our simulation data.

To account for the high friction of real skateboard decks, we apply DR during training, varying friction parameters to improve sim-to-real transfer of foot-mounting and body-alignment behaviors.

\begin{figure}[t]
    \centering
    \includegraphics[width=1.0\linewidth]{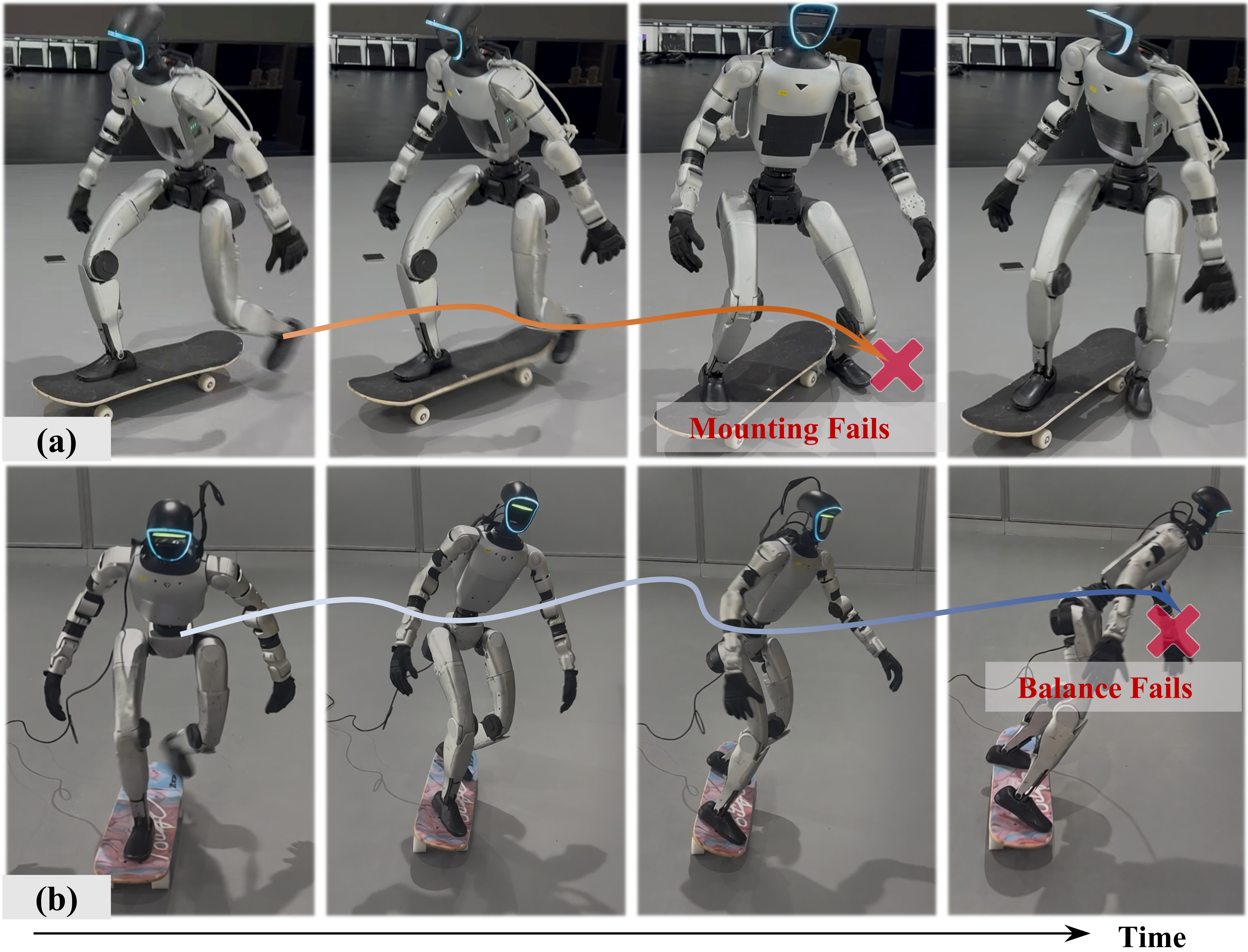}
    \caption{\textbf{Effect of Skateboard Physical Identification.} 
(a) Applying the compliant board parameters to the stiff board prevents mounting, as the real stiff board does not tilt under the robot’s step.
(b) Applying the stiff board parameters to the compliant board causes over-leaning and loss of stability during steering.}
    \label{fig:real}
\end{figure}
\subsubsection{Role of System Identification}

We evaluate the importance of skateboard system identification by focusing on the board’s tilt compliance, which determines how readily the board responds to humanoid leaning. Experiments are conducted using two skateboards with distinct tilt stiffness: one stiffer (black) and one more compliant (pink). As shown in Fig.~\ref{fig:cover} (a) and (b), when using the identified parameters corresponding to each physical board, the robot successfully performs push-and-steer maneuvers and achieves smooth board mounting and dismounting.

Cross-applying the identified parameters reveals the limitations of mismatched system models. In Fig.~\ref{fig:real} (a), when using parameters identified from the compliant board on the stiffer board, the robot fails to mount. In simulation, the board tilts under the robot’s stepping motion, allowing the policy to exploit this compliance for mounting, whereas the real stiff board remains nearly flat, breaking this assumption and preventing successful mounting. Conversely, in Fig.~\ref{fig:real} (b), applying the stiff-board parameters to the compliant board causes excessive leaning and loss of stability during steering, since the policy is not trained for such compliant dynamics. These results highlight that accurate identification of board tilt compliance is critical for sim-to-real transfer of humanoid skateboarding policies.

\section{Conclusion and Future Work}

In this work, we present \ours{}, a physics-aware learning framework that enables humanoid robots to perform agile skateboarding with active propulsion and lean-to-steer control. By modeling the coupled humanoid–skateboard dynamics and explicitly capturing the relationship between board tilt and truck steering, we decompose the task into pushing, steering, and transition phases and develop phase-specific learning strategies. We further introduce a trajectory-guided transition mechanism to enable smooth and reliable phase switching, which is critical for long-horizon stability. Extensive simulation and real-world experiments demonstrate that \ours{} enables continuous and robust skateboarding behaviors on the Unitree G1 humanoid robot platform.

We emphasize that HUSKY is a principled paradigm with utility beyond skateboarding. Its core mechanisms—kinematic coupling for underactuated tools and transitions between contact topologies—extend to broader tasks. Despite strong simulation and real-world performance, several limitations remain and warrant future investigation.

\textbf{Perception Integration.} The limited camera field of view prevents reliable observation of the board and wheel–ground interactions, restricting perception-driven feedback in the control loop. Incorporating visual state estimation is important for enabling perception-aware skateboarding control. 

\textbf{Complex Terrains.} Current experiments are conducted on relatively simple terrains, whereas human skateboarders routinely perform in complex environments such as skateparks while executing acrobatic maneuvers. Extending our framework to such scenarios will require richer motion priors and terrain-adaptive control strategies.

\section*{Acknowledgments}
Our training pipeline builds on mjlab~\cite{zakka2026mjlablightweightframeworkgpuaccelerated}, we gratefully acknowledge the open-source communities and contributors behind the frameworks, and thank Unitree Robotics~\cite{unitree-g1} for their support of the humanoid hardware used in this work.

\bibliographystyle{plainnat}
\bibliography{references}

\clearpage

\appendix

\subsection{Derivation of Skateboard Equality Constraint} 
\label{sec:app-proof}

\begin{figure}[t]
    \centering
    \includegraphics[width=1.0\linewidth]{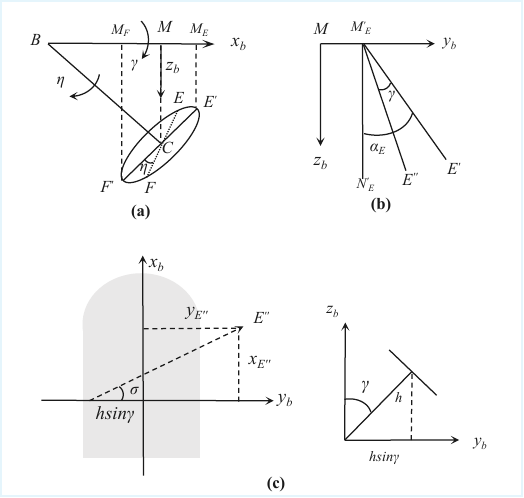}
    \caption{\textbf{Kinematic geometry of a skateboard truck.}  (a) Rotating about the truck pivot axis. 
        (b) Geometric construction for $\gamma$-rotation. 
        (c) Definition of steering angle $\sigma$ from top view.}
    \label{fig:app-proof}
\end{figure}

This appendix derives the kinematic relationship between the deck tilt angle $\gamma$ and the wheel axle steering angle $\sigma$ for a skateboard truck with fixed rake angle $\lambda$. The derivation follows a two-stage rotation sequence constrained by the wheel-ground contact condition~\cite{varszegi2015skateboard}.
In Fig.~\ref{fig:app-proof} (a), place the origin at the deck center projection $M = (0,0,0)$ with the $xy$-plane coinciding with the ground plane. Define:
\begin{itemize}
    \item Truck pivot center: $C = (0,0,h)$ where $h$ is the truck height
    \item Initial wheel positions: $E = (0,w,h)$, $F = (0,-w,h)$ where $2w$ is the truck width
    \item Kingpin axis $BC$ lies in the horizontal plane at fixed angle $\lambda$ relative to the $x$-axis
\end{itemize}

\textbf{$\eta$-rotation about kingpin axis $BC$.} Transforms wheel positions to:
\begin{align}
    E' &= \bigl(w\sin\eta\sin\lambda,\; w\cos\eta,\; h - w\sin\eta\cos\lambda\bigr) \\
    F' &= \bigl(-w\sin\eta\sin\lambda,\; -w\cos\eta,\; h + w\sin\eta\cos\lambda\bigr)
\end{align}
    
\textbf{$\gamma$-rotation about $x$-axis.} 
The board tilt corresponds to a rotation by angle $\gamma$ about the $x$-axis. To rigorously derive the coordinate transformation, we introduce auxiliary points that exploit the geometry of rotation about a coordinate axis (Fig.~\ref{fig:app-proof} (b)).
For wheel $E'$, we define:
\begin{itemize}
    \item $M_{E'} = (x_{E'}, 0, 0)$: orthogonal projection of $E'$ onto the $X$-axis
    \item The segment $M_{E'}\text{-}E'$ lies in the plane $x = x_{E'}$, perpendicular to the rotation axis
    \item Rotation about the $x$-axis preserves $x$-coordinates and moves $E'$ along a circular arc centered at $M_{E'}$ with radius $r_E = \|M_{E'}\text{-}E'\|$
\end{itemize}
Let $\alpha_E$ denote the angle between segment $M_{E'}\text{-}E'$ and the positive $z$-axis. By construction:
\begin{equation}
    r_E \sin\alpha_E = y_{E'}, \qquad r_E \cos\alpha_E = z_{E'}
    \label{eq:alpha_def}
\end{equation}
After rotation by $\gamma$, the angle between $M_{E'}\text{-}E''$ and the $z$-axis becomes $\alpha_E - \gamma$. Applying trigonometric addition formulas:
\begin{align}
    y_{E''} &= r_E \sin(\alpha_E - \gamma) 
            = y_{E'}\cos\gamma - z_{E'}\sin\gamma \\
    z_{E''} &= r_E \cos(\alpha_E - \gamma) 
            = z_{E'}\cos\gamma + y_{E'}\sin\gamma
\end{align}
with $x_{E''} = x_{E'}$ preserved by the rotation symmetry. Analogous definitions for $F'$ yield:
\begin{align}
    x_{F''} &= x_{F'} \\
    y_{F''} &= y_{F'}\cos\gamma - z_{F'}\sin\gamma \\
    z_{F''} &= z_{F'}\cos\gamma + y_{F'}\sin\gamma
\end{align}
Substituting the expressions for $E'$ and $F'$ gives the complete transformed coordinates.

\textbf{Wheel-Ground Contact Constraint.}
For both wheels to maintain ground contact simultaneously, their $z$-coordinates must be equal ($z_{E''} = z_{F''}$):
\begin{align}
    (h - w\sin\eta\cos\lambda)\cos\gamma + w\cos\eta\sin\gamma \\
    = (h + w\sin\eta\cos\lambda)\cos\gamma - w\cos\eta\sin\gamma
\end{align}
Simplifying yields the fundamental constraint:
\begin{equation}
    \cot\eta = \cos\lambda\cot\gamma
    \label{eq:cot_eta}
\end{equation}

\textbf{Steering Angle Definition.}
In Fig.~\ref{fig:app-proof} (c), the steering angle $\sigma$ is defined geometrically from the projection of $E''$ onto the $xy$-plane relative to the rotated truck center $C'' = (0,-h\sin\gamma,h\cos\gamma)$:
\begin{equation}
    \tan\sigma = \frac{x_{E''}}{y_{E''} + h\sin\gamma}
\end{equation}
which simplifies to:
\begin{equation}
    \tan\sigma\,(\cos\eta\cos\gamma + \sin\eta\cos\lambda\sin\gamma) = \sin\eta\sin\lambda
    \label{eq:sigma_def}
\end{equation}

Substituting Eq.~\eqref{eq:cot_eta} into Eq.~\eqref{eq:sigma_def}:
\begin{align}
    \tan\sigma\cos\lambda\left(\frac{\cos^2\gamma + \sin^2\gamma}{\sin\gamma}\right) &= \sin\lambda
\end{align}
Finally it yields the exact kinematic relationship, as illustrated in Eq~\eqref{eq:angle}:
\begin{equation}
    \tan\sigma = \tan\lambda \cdot \sin\gamma
    \label{eq:exact_relation}
\end{equation}
We embed this equivalent relationship into simulation using two implementations: (i) enforcing the constraint via MuJoCo equality constraints~\cite{todorov2012mujoco}, and (ii) directly setting the corresponding steer angle as a function of the tilt angle within the simulation environment. Both implementations produce similar kinematic behaviors in practice.

\begin{table}[ht]
    \caption{Skateboard Bodies}
    \label{tab:board-body}
    \centering
    \renewcommand{\arraystretch}{1.2}
    {\fontsize{8}{8}\selectfont
    
    \begin{tabular}{llll}
    \toprule
        {Name} & {Position} & {Type} & {Size} \\ \midrule
        Skateboard deck & 0 0 0 & Box &  0.8 0.2 0.02 \\ 
        Front truck & 0 0 -0.09 & Box  & 0.1 0.02 0.02\\ 
        Front left wheel & 0 0.07 0 & Cylinder  & 0.03 0.02 \\
        Front right wheel & 0 -0.07 0 & Cylinder  & 0.03 0.02 \\ 
        Rear truck  & 0 0 -0.09 & Box  & 0.1 0.02 0.02 \\
        Rear left wheel & 0 0.07 0 & Cylinder  & 0.03 0.02 \\
        Rear right wheel & 0 -0.07 0 & Cylinder  & 0.03 0.02 \\ 
    \bottomrule
    \end{tabular}}
\end{table}

\begin{table}[ht]
    \caption{Skateboard Joints}
    \label{tab:board-joint}
    \centering
    \renewcommand{\arraystretch}{1.2}
    {\fontsize{8}{8}\selectfont
    
    \begin{tabular}{lllll}
    \toprule
        {Name} & {Type} & {Joint Axis} & {Range} \\ \midrule
        Board tilt joint & Hinge & -1 0 0 & (-0.2, 0.2) \\ 
        Front truck joint & Hinge & 0 0 1 & (-0.1, 0.1) \\ 
        Rear truck joint  & Hinge & 0 0 -1 & (-0.1, 0.1) \\ 
        Wheel joints & Hinge & 0 1 0 & Continuous \\ 
    \bottomrule
    \end{tabular}}
\end{table}

\subsection{Skateboard Model}
\begin{figure}[t]
    \centering
    \includegraphics[width=0.9\linewidth]{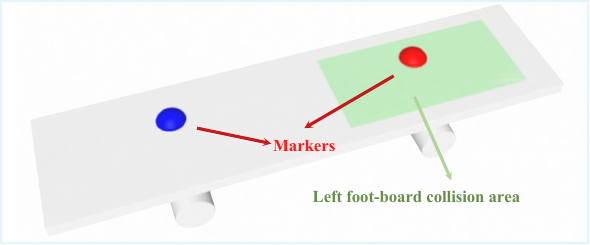}
    \caption{\textbf{Skateboard Model in Training.}  Red and blue markers indicate stabel foot placement points, while thelight green areas denote collision zones used for foot-board collision detection.}
    \label{fig:board-model}
\end{figure}

The key components of our simplified skateboard model in MuJoCo are summarized in Table~\ref{tab:board-body} and Table~\ref{tab:board-joint}. In simulation, the wheel–ground contact is modeled with six dimensions (\textit{condim} = 6), including two tangential, one torsional, and two rolling directions, to realistically capture wheel–ground interactions. The rolling friction coefficient is set to 0.001 to reflect low-resistance wheel motion. As shown in Fig.~\ref{fig:board-model}, two virtual markers are positioned above the trucks to guide stable foot contacts, as illustrated by the blue and red spheres.

\label{sec:app-model}

\begin{table*}[t]
    \caption{Task Reward terms and weights.}
    \label{tab:reward_desgin}
    
    \centering
    {
    \begin{tabular}{llll}
\toprule
Term & Expression & Weight & Meaning \\
\midrule
\rowcolor[HTML]{EFEFEF} \textbf{Pushing Phase} $r_t^{\text{push}}$ & & & \\
\midrule
Linear velocity tracking &
$\exp(-|  v_{\rm board} -  v_{\rm cmd} |^2 / \sigma_v^2 )$
& 3.0 & Track commanded forward velocity \\

Yaw alignment &
$\exp(- \lvert \psi_{\rm robot} - \psi_{\rm board} \rvert^2 / \sigma_{\rm yaw}^2 )$
& 1.0 & Align robot yaw with skateboard during pushing \\

Feet air time &
$\mathbb{I}(T_{\text{air}}^{\rm left\_foot} \in [T_{\text{air}}^{\min}, T_{\text{air}}^{\max}]) \cdot \mathbb{I}( v_{\rm cmd} >  v_{\rm th})$
& 3.0 & Encourage proper left-foot lift timing during pushing \\

Ankle parallel~\cite{hr:huang2025learning} &
$\mathbb{I}(\mathrm{Var}(z_{\rm left\_ankle}) < 0.05 \ )\cdot \mathbb{I}(\text{left foot on ground}))$
& 0.5 & Encourage left foot to remain parallel during pushing \\

AMP style reward &
$\alpha \cdot \max\left(0,\ 1 - {1}/{4}(d - 1)^2\right)$
& 5.0 & Encourage human-like natural pushing behavior \\

\midrule
\rowcolor[HTML]{EFEFEF} \textbf{Steering Phase} $r_t^{\text{steer}}$& & & \\
\midrule
Steer feet contact &
$2 *\, \mathbb{I}(\text{both feet on board}) - \mathbb{I}(\text{left foot on ground})$
& 3.0 & Encourage both feet on board and avoid ground contact \\

Joint position deviation &
$\exp(- \lVert \bm{\theta}_t - \hat{\bm{\theta}}_t \rVert_2^2 / \sigma_{\rm jpos}^2)$
& 1.5 & Maintain nominal humanoid steering pose \\

Heading tracking &
$\exp(- (\psi_{\rm board} - \psi)^2 / \sigma_\psi^2)$
& 5.0 & Track desired board direction \\

Board tilt tracking &
$\exp(- (\gamma - \gamma_{\rm ref})^2 / \sigma_\gamma^2)$
& 4.0 & Align board lean with physics-guided reference \\

Feet marker distance &
$\exp(- \lVert \bm{p}_{\rm foot} - \bm{p}_{\rm marker} \rVert_2^2 / \sigma_m^2)$
& 1.0 & Encourage feet near preferred foot markers \\

\midrule
\rowcolor[HTML]{EFEFEF} \textbf{Transition} $r_t^{\text{trans}}$& & & \\
\midrule
Keybody position tracking &
$\exp(- \lVert \bm{p}^{\mathcal{K}}_t - \hat{\bm{p}}^{\mathcal{K}}_t \rVert_2^2 / \sigma_{\rm pos}^2)$
& 10.0 & Follow trajectory-planned keybody positions \\

Keybody orientation tracking &
$\exp(- \lVert \bm{q}^{\mathcal{K}}_t \ominus \bm{q}^{\mathcal{K}}_t \rVert_2^2 / \sigma_{\rm rot}^2)$
& 10.0 & Follow trajectory-planned keybody orientations \\

\midrule
\rowcolor[HTML]{EFEFEF} \textbf{Regularization} $r_t^{\text{reg}}$& & & \\
\midrule
Skateboard wheel contact &
$\mathbb{I}\big(\sum_{i=1}^{4} c_i = 4 \big)$
& 0.5 & Reward full wheel contact, avoid unrealistic detachment \\

Joint position limits &
$\mathbb{I}(\bm{q}_t \notin [\bm{q}_{\rm min}, \bm{q}_{\rm max}])$
& -5.0 & Keep joints within safe limits \\

Joint velocity &
$\lVert \dot{\bm{q}}_t \rVert_2^2$
& -1e-3 & Penalize high joint speeds \\

Joint acceleration &
$\lVert \ddot{\bm{q}}_t \rVert_2^2$
& -2.5e-7 & Penalize abrupt joint accelerations \\

Joint torque &
$\lVert \bm{\tau}_t \rVert_2^2$
& -1e-6 & Penalize excessive torque \\

Action rate &
$\lVert \bm{a}_t - \bm{a}_{t-1} \rVert_2^2$
& -0.1 & Encourage smooth actions \\

Action smoothness &
$\lVert \bm{a}_t - 2 \bm{a}_{t-1} + \bm{a}_{t-2} \rVert_2^2$
& -0.1 & Reduce oscillations in control commands \\

Collision & 
$\mathbb{I}_{\rm collision}$ 
& -10.0 & Penalize self-collisions \\

\bottomrule
\end{tabular}

    }
\end{table*}

\subsection{Reward Functions} 
\label{sec:app-reward}
The reward terms, corresponding to pushing, steering, transitions, and regularization, are summarized in Table~\ref{tab:reward_desgin}.

\begin{table}[ht]
    \centering
    {
        \caption{Hyperparameters related to PPO.}
    \label{tab:ppo_params}
    \begin{tabular}{lc}
        \toprule
        Hyperparameter & Value \\
        \midrule
        Optimizer & Adam \\
        Batch size & 4096 \\
        Mini Batches & 4 \\
        Learning epoches & 5 \\
        Entropy coefficient & 0.005 \\
        Value loss coefficient & 1.0 \\
        Clip param & 0.2 \\
        Max grad norm & 1.0 \\
        Init noise std & 1.0 \\
        Learning rate & 1e-3 \\
        Desired KL & 0.01 \\ 
        GAE decay factor($\lambda$) & 0.95\\
        GAE discount factor($\gamma$) & 0.99\\
        Actor MLP size & [512, 256, 128] \\
        Critic MLP size & [512, 256, 128] \\
        MLP Activation & ELU \\
        \bottomrule
    \end{tabular}
}
\end{table}

\subsection{Training Details} 
\label{sec:app-train}
We define a skateboarding cycle of 6 seconds with a hybrid phase structure, including a pushing phase (40\%), a foot-mounting transition phase (10\%), a steering phase (45\%), and a dismounting transition phase (5\%). The phase variable $\Phi \in [0,1)$ is divided into Pushing $[0,0.4]$, Mounting $[0.4,0.5]$, Steering $[0.5,0.95]$, and Dismounting $[0.95,1.0)$.  All parallel
envs initialize from Pushing, with rewards computed according
to the active phase. The RL policies are trained on an NVIDIA RTX 5080 server, with each iteration taking approximately 2–3 seconds. The tracking-based baseline follows DeepMimic~\cite{peng2018deepmimic} style, while the gait-based uses a periodic gait clock~\cite{TienKungLab2024}. Training a policy for real-world deployment requires roughly 20 hours in total. The PPO hyperparameters used are summarized in Table~\ref{tab:ppo_params}.

\subsection{Foot Contact Detection} 
To accurately detect foot contacts with the ground and skateboard, we employ separate contact sensors for each foot–surface pair: left foot–ground, right foot–ground, left foot–board, and right foot–board. These sensors filter the corresponding contact interactions to provide reliable feedback.

In addition, to prevent foot penetration during training, small contact collision regions are placed on the skateboard at the designated foot placement locations. These regions ensure that only intended foot–board contacts are registered, preventing the policy from exploiting unintended collisions with other parts of the skateboard to obtain contact rewards. As shown in Fig.~\ref{fig:board-model}, the active collision regions are highlighted as light green shaded areas on the board.

\begin{figure}[t]
    \centering
    \includegraphics[width=1.0\linewidth]{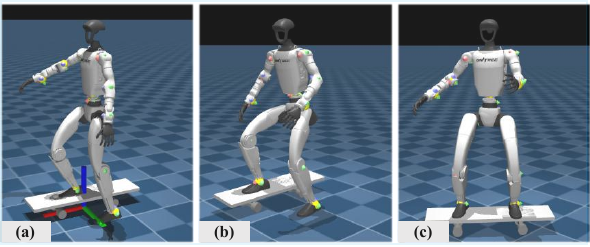}
    \caption{\textbf{Key body reference poses.} 
    (a) Default pose used for policy training. 
    (b) Reference pose for the pushing phase. 
    (c) Reference pose for the steering phase. 
    All poses are annotated from the human motion dataset $\mathcal{M}$ to ensure realistic humanoid posture configuration.}
    \label{fig:trans_pose}
\end{figure}

\subsection{Reference Pose} 
The initial pose used at the start of training is shown in Fig.~\ref{fig:trans_pose} (a), while the canonical reference poses for the pushing and steering phases are shown in Fig.~\ref{fig:trans_pose} (b) and Fig.~\ref{fig:trans_pose} (c). All these poses are annotated from the human motion dataset $\mathcal{M}$ to ensure realistic posture and limb configurations.

The key bodies used to define these poses include the pelvis, torso, left and right hips, knees, and ankles, as well as the left and right shoulders, elbows, and wrists. Representing the humanoid with these key bodies provides a compact yet expressive description of its configuration, facilitating smooth and effective transitions between phases.

\subsection{Evaluations on More Terrains}
\begin{figure}[t]
    \centering
    \includegraphics[width=\linewidth]{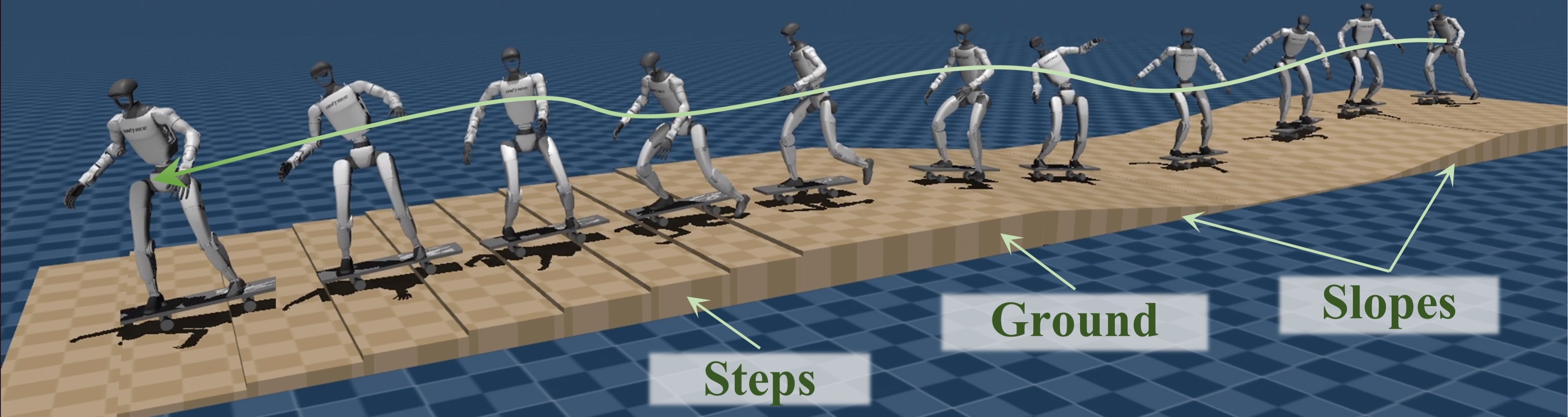}
    \caption{\textbf{Visualizations on more terrains.} HUSKY maintains robust zero-shot performance on unseen terrains including slopes and steps. 
}
    \label{fig:terrain}
\end{figure}

While our experiments were conducted on flat ground, the real-world indoor and outdoor testing environments already contained tile gaps and varying frictions. Our heavy DR strategy ensures that the policy manages these unmodeled surface irregularities. We evaluated HUSKY on inclined slopes ($5^\circ$--$10^\circ$) and steps ($3-6cm$) (Fig.~\ref{fig:terrain}), with results in Table~\ref{tab:exp_terrain} confirming its adaptability to uneven terrains. We acknowledge that extreme scenarios, such as vertical ramps or bowl parks, pose significant risks for humanoids and require vision-based planning beyond the current scope. 

\begin{table}[t]
\centering
\caption{Evaluations on More Terrains}
\captionsetup{singlelinecheck=false,justification=raggedright}
\label{tab:exp_terrain}
\begin{tabular}{lccc}
\toprule
\textbf{Terrain Settings} & $E_{\text{succ}}\!\uparrow$ & $E_{\text{smth}}\!\downarrow$ & $E_{\text{contact}}\!\downarrow$ \\
\midrule

HUSKY-Slope & \phantom{} ${{95.02}}\!\pm\! \mathsmaller{0.14}$ & ${0.038}\!\pm\! \mathsmaller{0.006}$ &  ${0.001}\!\pm\! \mathsmaller{0.001}$ \\

HUSKY-Steps & \phantom{} ${{89.21}}\!\pm\! \mathsmaller{0.25}$ & ${0.042}\!\pm\! \mathsmaller{0.009}$ &  ${0.003}\!\pm\! \mathsmaller{0.002}$\\

HUSKY-Flat Ground & ${{100.00}}\!\pm\! \mathsmaller{0.00}$ & ${0.033}\!\pm\! \mathsmaller{0.005}$ &  ${0.001}\!\pm\! \mathsmaller{0.001}$ \\

\bottomrule
\end{tabular}
\end{table}

\subsection{Results of Skateboard Identification} 
The gains of the PD controller were determined via system identification for two skateboards. The moment of inertia ($I$) of each plant was calculated using the formula for a rectangular prism, resulting in $I = 7.15 \times 10^{-3}~\mathrm{kg\cdot m^2}$ for the first skateboard and $I = 8.70 \times 10^{-3}~\mathrm{kg\cdot m^2}$ for the second.

Analysis of the experimental decaying oscillation responses confirmed that both systems exhibited underdamped second-order dynamics. For the first skateboard, an oscillation period of $T = 0.107~\mathrm{s}$ was observed, with successive peak amplitudes of $\theta(t) = 0.614$ and $\theta(t+T) = 0.0108$. This yielded controller gains of $K_p = 34.835$ and $K_d = 0.540$. The second skateboard demonstrated a period of $T = 0.185~\mathrm{s}$, with peak amplitudes of $\theta(t) = 0.583$ and $\theta(t+T) = 0.0081$, which resulted in gains of $K_p = 14.677$ and $K_d = 0.402$.

\begin{figure}[t]
    \centering
    \includegraphics[width=\linewidth]{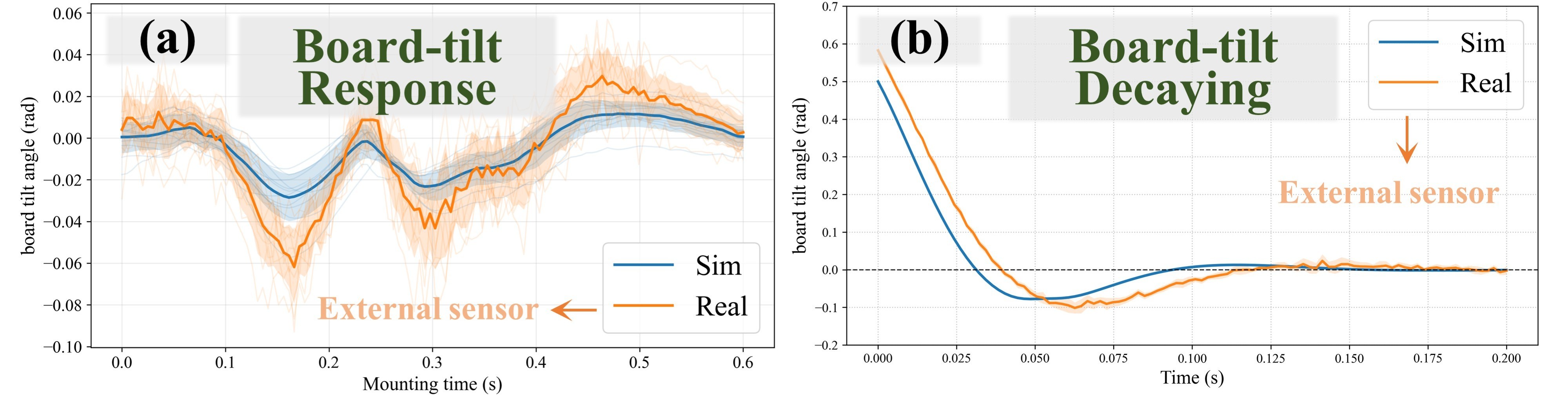}
    \caption{\textbf{Board-tilt response between sim and real.} The board-tilt response curves during mounting and the board-tilt decaying results demonstrate the physics fidelity of our skateboard modeling approach.
}
    \label{fig:sys}
\end{figure}

We model skateboard suspension via a lean-to-steer coupling using an equivalent spring–damper model. While abstracting detailed bushing deformation, it captures the dominant kinematic relationship within a $15^\circ$ roll range, which is critical for steering. We provide two-fold evidence to support physics fidelity. (1) A board-tilt response (Fig.~\ref{fig:sys} (a)) during mounting validates transient dynamics. A free-decay roll response (Fig.~\ref{fig:sys} (b)) shows close alignment between hardware and the identified model in decaying oscillation; (2) We clarify although a flat deck is used, spatial guidance and foot markers ensure contact remains within the stable central region, avoiding unmodeled kicktails. And DR of CoM offsets and surface friction mitigates sensitivity to simplified geometry.

\subsection{Effects of Transition Design} 

\begin{figure}[h]
    \centering
    \includegraphics[width=\linewidth]{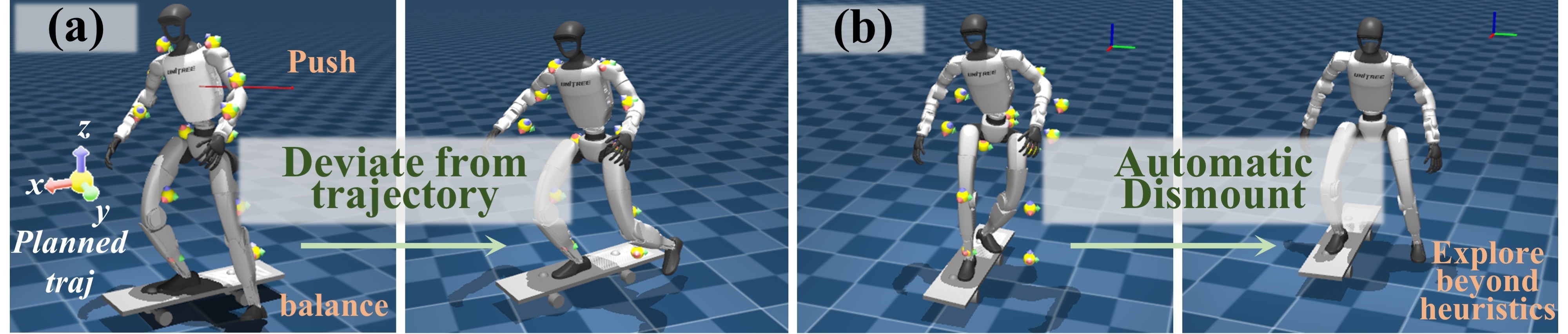}
    \caption{\textbf{Humanoids explore beyond heuristic transition.} Humanoid deviates from the planned trajectories proves the policy mastered the underlying balancing physics instead of rigidly following heuristics.
}
    \label{fig:trans}
\end{figure}

Our key insight is leveraging \textit{explicit guidance} as a minimal prior to prune the search space, not as a hard-coded constraint. In Table~\ref{tab:sens_trans}, HUSKY maintains high performance even under Canonical Pose Perturbation ($\pm 5cm$), Transition Duration Scaling ($\pm 20\%$), and Interpolation Variants (Spline), showing that the policy is not overfitted to specific trajectories. Under external forces, the agent prioritizes stability over following trajectories (Fig.~\ref{fig:trans}). This ability to deviate from the heuristics proves the policy mastered the underlying balancing physics, offering a scalable approach for multi-phase tasks.

\begin{table}[h]
\centering
\caption{Sensitivity Analysis of Transition}
\captionsetup{singlelinecheck=false,justification=raggedright}
\label{tab:sens_trans}
\begin{tabular}{lccc}
\toprule
\textbf{Transition Settings} & $E_{\text{succ}}\!\uparrow$ & $E_{\text{smth}}\!\downarrow$ & $E_{\text{contact}}\!\downarrow$ \\
\midrule

Pose-Perturbed & \phantom{} ${{99.63}}\!\pm\! \mathsmaller{0.12}$ & ${0.037}\!\pm\! \mathsmaller{0.007}$ &  ${0.002}\!\pm\! \mathsmaller{0.001}$ \\

Time-Scaled & ${{100.00}}\!\pm\! \mathsmaller{0.00}$ & ${0.035}\!\pm\! \mathsmaller{0.003}$ &  ${0.001}\!\pm\! \mathsmaller{0.002}$\\

Path-Varied  & \phantom{} ${{98.54}}\!\pm\! \mathsmaller{0.93}$ & ${0.033}\!\pm\! \mathsmaller{0.004}$ &  ${0.002}\!\pm\! \mathsmaller{0.002}$ \\

HUSKY & ${{100.00}}\!\pm\! \mathsmaller{0.00}$ & ${0.033}\!\pm\! \mathsmaller{0.005}$ &  ${0.001}\!\pm\! \mathsmaller{0.001}$ \\

\bottomrule
\end{tabular}
\end{table}

\subsection{More Skateboards} 
\begin{figure}[t]
    \centering
    \includegraphics[width=1.0\linewidth]{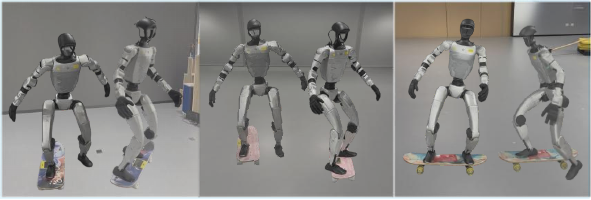}
    \caption{\textbf{Humanoid Skateboarding on Diverse Boards.}  \ours{} enables the humanoid to perform skateboarding behaviors across multiple skateboard platforms, demonstrating the generality and robustness of our method.}
    \label{fig:more-board}
\end{figure}

We further evaluate \ours{} on a variety of skateboard platforms. Results show that the humanoid successfully performs skateboarding behaviors, including pushing, steering, and phase transitions. As shown in Fig.~\ref{fig:more-board}, these results confirm the generality and effectiveness of our approach.

\subsection{More Findings} 
We conducted comparative experiments across different simulation frameworks. While IsaacLab~\cite{mittal2025isaac} offers superior parallelization, policies trained within this environment frequently failed during sim-to-sim validation due to excessive foot-skateboard slippage. These issues were resolved upon integrating mjlab~\cite{zakka2026mjlablightweightframeworkgpuaccelerated} for fine-grained simulation. The success of mjlab deployment suggests that its underlying physics engine provides a more accurate representation of the non-linear friction and contact dynamics crucial for skating. 

\end{document}